\documentclass[11pt]{article} 

\usepackage[utf8]{inputenc} 

\usepackage{geometry} 
\usepackage{graphicx} 

\usepackage{booktabs} 
\usepackage{array} 
\usepackage{paralist} 
\usepackage{verbatim} 
\usepackage{subfig} 

\usepackage{fancyhdr} 
\usepackage{sectsty}
\allsectionsfont{\sffamily\mdseries\upshape} 

\usepackage[nottoc,notlof,notlot]{tocbibind} 
\usepackage[titles,subfigure]{tocloft} 

\title{Oversight of Unsafe Systems via Dynamic Safety Envelopes}
\author{David Manheim}

\begin{document}
\maketitle
\abstract{This paper reviews the reasons that Human-in-the-Loop is both critical for preventing widely-understood failure modes for machine learning, and not a practical solution. Following this, we review two current heuristic methods for addressing this. The first is provable safety envelopes, which are possible only when the dynamics of the system are fully known, but can be useful safety guarantees when optimal behavior is based on machine learning with poorly-understood safety characteristics. The second is the simpler circuit breaker model, which can forestall or prevent catastrophic outcomes by stopping the system, without any specific model of the system. This paper proposes using heuristic, dynamic safety envelopes, which are a plausible half--way point between these approaches that allows human oversight without some of the more difficult problems faced by Human-in-the-Loop systems. Finally, the paper concludes with how this approach can be used for governance of systems when otherwise unsafe machine learning is deployed.}\\

There are important shortcomings of fully autonomous systems that can be remedied by allowing a human to control certain parts of the system, known as Human-in-the-Loop. As noted by Richard Danzig\cite{Danzig2018} among others, there are fundamental dynamics that make this solution unworkable.  One of the issues is that machine learning technology runs on computers that are orders of magnitudes faster than humans. Not only that, but they are often far less expensive, make better use of large datasets, and can react in real time. These factors are not incidental to the use of AI, but rather form a large part of the reason that (narrow) AI is used. For this reasons, human interruption of AI decision making often obviates the very reason it is being used.

As discussed by Paul et al.\cite{Paul2018a}, for the purposes of autonomous systems, we can consider three levels of automation: direct automation of systems that automatically execute pre-defined actions, machine learning and narrow AI that can act autonomously given a narrow and rigidly defined domain, and general AI that can make decisions in a way that is context aware and reflects understanding of the domain. While there are clear drawbacks to simple automation, and clear dangers from mis-aligned or very capable general intelligence\cite{Yudkowsky2016b}, the limitations of the second of these three levels is the focus of the present discussion, and it will be referred to using the somewhat ambiguous term ``machine learning.''

Given that we are both ruling out automation as insufficient, and strong artificial intelligence as, at the very least, beyond our current capabilities, the oft-repeated suggestion of ``Human-in-the-Loop'' to prevent failures seem naive - if we wanted to slow the process down to human speed, and require humans to be involved, there is little reason for AI rather than automation of pre-defined choices for the human decision maker. If the interactions are too fast for true human control, on the other hand, as Danzig colorfully put it, ``Human decisionmakers are riders traveling across obscured terrain with little or no ability to assess the powerful beasts that carry and guide them.'' This rules out what has been called the Licklider's augmentation approach\cite{Schwartz2018} which was temporarily dominant in chess\cite{Kasparov2010}, for enabling cooperative use of AI, at least when real-time interaction is needed. At the same time, failure modes of machine learning, including vulnerability to adversarial examples, lack of awareness of changing context, and others, necessitate a form of human involvement\footnote{It is sometimes argued that these flaws can be remedied with better systems. This is true, but only when ``better'' refers to contextually aware, generally intelligent systems - that is, General Artificial Intelligence.}. 

There are a few ways that we can square the conceptual circle of needing impossibly fast and accurate human oversight, which range from provably impossible, to currently unreachable, to merely hard. The first is to build systems that are not susceptible to the various ways in which agents in complex multi-party systems can fail (as outlined by Manheim\cite{Manheim2018},) is provably impossible in at least some cases, such as preventing adversarial examples, as proved by Shahafi et al.\cite{Shafahi2018}. The second, which currently resides squarely in the realm of science fiction, are what Robin Hanson has christened ``Ems,'' or uploaded human intelligences that can be run on a computational substrate\cite{Hanson2016}. These could conceivably be run fast enough to allow the type of real-time collaboration that Licklider envisioned, but are well beyond our current technological reach. The final three, however, are more realistic paradigms.

The first of these potential solutions is the ``Provable Limited Scope'' model. This is exemplified by the Responsibility-Sensitive Safety (RSS) model of Shalev-Shwartz, Shammah, and Shashua\cite{Shalev-Shwartz2017}. In this model, the autonomous systems are free to build whatever driving policy they wish, with the restriction that the car stays within a safety envelope defined via specific restrictions on its position and speed relative to other cars. For example, it muse stay far enough from the car in front that it can stop without impacting even if that car stops at some defined physically feasible maximum deceleration. In the case of RSS, it is possible to prove that a system of agents abiding by these rules will not ever collide. If such a model can be constructed for a given context, it is reasonable to give a machine learning system free range with the knowledge that it will act safely. 

A potentially simpler, but much less effective version of this is the second possibility, ``Heuristic Limited Scope'' which provides a far weaker guarantee. As a simple example of this model, the ``circuit breaker'' for financial markets sets limits on the maximum price movement of financial assets. By relying on a purely heuristic barrier that does not relate to the dynamics of the system in question, this provides much less assurance that the system is safe, as should be obvious from the existence of specific ongoing failure modes used as examples of multi-agent failure modes, per Manheim\cite{Manheim2018}, such as algorithmic momentum ignition\cite{shorter2014high}.

The final method relies on contextual awareness and limits provided dynamically. This falls short of the provable guarantee of the first model, but can be far less limited than the heuristic model. This ``Dynamic Safety Envelope'' would rely on automated systems for detecting and understanding the current system, potentially combined with human oversight, and providing parameters for safety. In this case, the envelope does not need to be redefined in real-time as decisions are made, nor does it need to be a fixed heuristic. Instead, changes can be specified at the potentially much slower speed needed for humans to orient to broader systemic changes.

For example, consider a system for automated detection of and response to automated propaganda, such as automated banning of accounts. This could be a significant liability if the automated system learns from live data, since a clever opponent might craft messages from systems that would correctly be flagged as propaganda bots, but which also are adversarial examples crafted to poison the classifier\cite{Li2016,Jagielski2018}, in this case to cause it to ban accounts chosen by the attacker. In this case, change-point detection algorithms can identify that there were changes in the distribution of the bot behavior, and if detected, could stop itself to await review, or simply exclude those data points from the training set until further review. (Other parts of the safety envelope might include limits on which types of accounts to suspend without human review, etc.) 

By limiting the need for human intervention to where the system notes unusual behavior, humans can decide whether the current issue needs further review, without needing to stay involved at each decision point. That review can be done both more leisurely, and via help from other algorithmic analyses. For example, the change in bot behavior might be compared to shifts found in a broader set of data sources via change point detection across multiples sources and streams, such as Mei's work \cite{Mei2010}. This might make it clear that the detected change is part of a broader shift, rather than an attack.

If these sorts of contextual, human assisted automated systems can be trained to provide approximations of a safe-envelope for decision making, they need not be controlled by the same actor as the one controlling the agents. Regulation of machine learning systems can be done by an industry consortium or a government regulator. The human factors and joint decisions obviously lead to drawbacks and limitations regarding what safety is desired, and the extent to which the safe envelope of behavior is conservative or liberal in definition. Still, this approach seems like a promising way forward in the short term for deployment of AI that is safer than currently envisioned, even though it is not fully safe. This class of system does not, to the knowledge of the author, currently exist, but seems to be a promising method of allowing human oversight without needing real-time oversight. This could lead to marginally safer systems even short of provable security, or even security with high-probability. Despite being essentially a stop-gap measure, it seems like a reasonable middle ground between restricting AI to provably safe domains, and allowing arbitrary AI to be deployed with the untenable assumption that nothing will go wrong.

\bibliographystyle{acm}

\begin{thebibliography}{10}
\bibitem{Danzig2018}
{\sc Danzig, R.}
\newblock {Technology Roulette: Managing Loss of Control as Many Militaries
  Pursue Technological Superiority}.
\newblock Tech. rep., Center for a New American Security, 2018.

\bibitem{Hanson2016}
{\sc Hanson, R.}
\newblock {\em {The Age of Em: Work, Love, and Life when Robots Rule the
  Earth}}.
\newblock Oxford University Press, 2016.

\bibitem{Jagielski2018}
{\sc Jagielski, M., Oprea, A., Biggio, B., Liu, C., Nita-Rotaru, C., and Li,
  B.}
\newblock {Manipulating machine learning: Poisoning attacks and countermeasures
  for regression learning}.
\newblock {\em arXiv preprint arXiv:1804.00308\/} (2018).

\bibitem{Kasparov2010}
{\sc Kasparov, G.}
\newblock {The chess master and the computer}.
\newblock {\em The New York Review of Books 57}, 2 (2010), 16--19.

\bibitem{Li2016}
{\sc Li, B., Wang, Y., Singh, A., and Vorobeychik, Y.}
\newblock {\em {Data Poisoning Attacks on Factorization-Based Collaborative
  Filtering}}.
\newblock aug 2016.

\bibitem{Manheim2018}
{\sc Manheim, D.}
\newblock {Overoptimization Failures and Specification Gaming in Multi-agent
  Systems}.
\newblock {\em arXiv preprint arXiv:1810.10862\/} (2018).

\bibitem{Mei2010}
{\sc Mei, Y.}
\newblock {Efficient scalable schemes for monitoring a large number of data
  streams}.
\newblock {\em Biometrika 97}, 2 (jun 2010), 419--433.

\bibitem{Paul2018a}
{\sc Paul, C., Clarke, C.~P., Triezenberg, B.~L., Manheim, D., and Wilson, B.}
\newblock {Automation, Machine Learning, and Computational Propaganda}.
\newblock In {\em Requirements for Better C2 and Situational Awareness of the
  Information Environment}. RAND Corporation, 2018, ch.~Appendix.

\bibitem{Schwartz2018}
{\sc Schwartz, O.}
\newblock {Competing Visions for AI}.
\newblock {\em Digital Culture {\&} Society 4}, 1 (2018), 87--106.

\bibitem{Shafahi2018}
{\sc Shafahi, A., Huang, W.~R., Studer, C., Feizi, S., and Goldstein, T.}
\newblock {Are adversarial examples inevitable?}
\newblock {\em arXiv preprint arXiv:1809.02104\/} (2018).

\bibitem{Shalev-Shwartz2017}
{\sc Shalev-Shwartz, S., Shammah, S., and Shashua, A.}
\newblock {On a formal model of safe and scalable self-driving cars}.
\newblock {\em arXiv preprint arXiv:1708.06374\/} (2017).

\bibitem{shorter2014high}
{\sc Shorter, G.~W., and Miller, R.~S.}
\newblock {\em {High-frequency trading: background, concerns, and regulatory
  developments}}, vol.~29.
\newblock Congressional Research Service Washington, DC, 2014.

\bibitem{Yudkowsky2016b}
{\sc Yudkowsky, E.}
\newblock {The AI Alignment Problem: Why It's Hard, and Where to Start}.
\newblock In {\em Symbolic Systems Distinguished Speaker\/} (2016), Stanford
  University.

\end{thebibliography}

\end{document}